# Hybrid Genetic Algorithm and Mixed Integer Linear Programming for Flying Sidekick TSP

André Rossi Kuroswiski[1], Humberto Baldessarini Pires[1], Angelo Passaro[2], Lamartine Nogueira Frutuoso Guimarães[2], Edson Luiz França Senne[3]

[1]Instituto Tecnológico de Aeronáutica (ITA), São José dos Campos/SP – Brasil
[2]Instituto de Estudos Avançados (IEAv), São José dos Campos/SP – Brasil
[3]Universidade Federal de São Paulo (UNIFESP), São José dos Campos/SP – Brasil

*Abstract* – The increasing use of drones to perform various tasks has motivated an exponential growth of research aimed at optimizing the use of these means, benefiting both military and civilian applications, including logistics delivery. In this sense, the combined use of trucks and drones has been explored with great interest by Operations Research. This work presents mathematical formulations in Mixed Integer Linear Programming and proposes a hybrid Genetic Algorithm (HGenFS) for optimizing a variation of the Traveling Salesman Problem (TSP) called Flying Sidekick TSP (FSTSP), in which truck and drone cooperate. The results obtained confirmed that the adopted formulation for the exact solution is suitable for solving problems up to ten customers, and the HGenFS proved to be capable of finding optimal solutions for the FSTSP in a few seconds by incorporating specific heuristics and a local search phase.

## I. INTRODUCTION

Historically, the use of drones for military purposes dates back to the first half of the last century [1]. However, with the technological advancement of these devices in the last two decades, the military application of drones has become more widespread and intense, and their use has been studied from various perspectives, including those related to the civilian area, such as logistics operations [2].

In the meantime, the delivery service of goods has expanded considerably, and due to high competition in this market niche, consumers choose companies based on the speed, flexibility, and shipping costs offered. This has motivated companies in the industry to implement technological innovations in delivery services, especially in the last mile delivery stage [3], a term used to define the transport of products from distribution centers to their final destination, such as a house or business [4].

The optimization of the delivery service in the last mile delivery has been explored because this stage has a significant impact on the profit of companies [5]. For example, many technology companies, such as Google and Amazon, as well as traditional logistics companies, such as UPS and FedEx, have conducted experiments using drones with the aim of reducing costs and providing a cheaper, more efficient, and faster mode of delivery [6].

The choice to use drones is justified by the fact that these devices are not restricted to the road network nor limited by terrain, unlike trucks, which are traditional vehicles used for this purpose [7], [8]. However, the disadvantages of drones refer to the limited cargo capacity and the flight time limited by the Battery [9].

Due to the inherent advantages of these two types of vehicles, the use of a truck-drone system comprises a combination that enhances the best characteristics of trucks and drones. Optimizing this process can make the delivery service more efficient and possibly cheaper. Consequently, this has led to different new logistics problems, arousing increasing interest from the area of Operations Research [10].

In this sense, this work proposes an approach to a variation of the Traveling Salesman Problem (TSP), introduced in [11], in which a truck and a drone operate collaboratively to carry out deliveries. The services are performed by exploiting the best characteristics of each vehicle: the high average speed of the drone and the large cargo capacity of the truck. This type of problem has broad application in the military area, where, for example, the customers served can be considered targets or objects of interest in an operational scenario, and support vehicles or mobile bases perform the function of the truck. For solving the problem, a mathematical formulation modeled through Mixed Integer Linear Programming (MILP) is presented, and a heuristic method is proposed in which hybrid Genetic Algorithms were used to achieve viable solutions.

## II. LITERATURE REVIEW

In recent years, more than three hundred articles have been published on optimization problems using trucks and drones, and they can be classified into different categories of truck-drone systems [12]: trucks as a means of support for drone operations; drones supporting delivery services carried out by trucks; drones and trucks in independent tasks; and drones and trucks operating synchronously.

In this work, the scope of the literature review will be restricted to the most promising research related to the category of synchronized truck-drone system. The use of drones in optimizing a delivery process was first proposed in [11]. This work explored a variation of the Traveling Salesman Problem (TSP), defined as FSTSP (Flying Sidekick TSP), in which a drone and a truck form a system that operates cooperatively.

In the FSTSP, each customer is served only once, either by the drone or by the truck, which is used to make deliveries that exceed the cargo capacity or the autonomy of the drone. In addition, each customer node can only be visited once. The drone can be launched from the depot or from the truck. Each flight is called a sortie and comprises the takeoff from a customer node on the truck for delivery to only one customer, as well as the reunion at a different customer node on the truck. Launches and landings must be made with the truck stopped, and intermediate landings to save battery power are not allowed. The truck and drone must wait for each other at a customer node on the truck. The objective function is to minimize the completion time of the routes [11].

Approaches like the FSTSP constitute an NP-hard problem from the perspective of computational complexity [13]. For this reason, some heuristic methods have been explored to solve this type of problem [14]. The one presented in [11] proposes a MILP formulation that was not able to optimize, in thirty minutes, the instances related to ten customers, and three heuristic methods based on K-Nearest Neighbor (KNN) and scanning algorithms were proposed.

Another problem that shares the main characteristics of the FSTSP is the so-called TSP with Drone (TSP D). In [15], the authors modeled the problem with Integer Linear Programming (ILP) and developed heuristics based on Local Search and Dynamic Programming, finding optimal solutions in up to sixty minutes for instances of up to fifteen customers.

In Figure 1, it is possible to visualize possible solutions for the classic TSP (a), the FSTSP (b), and the TSP-D (c).

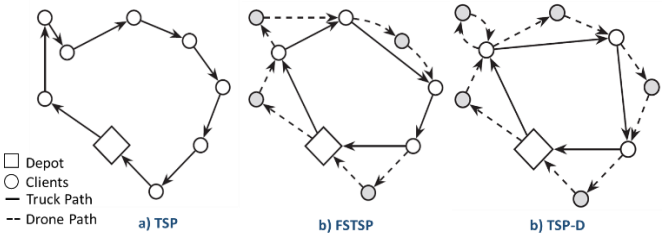

Fig. 1. TSP solutions comparison - (a), FSTSP (b) e TSP-D (c). Source: [10].

The main differences between the TSP-D and the FSTSP are: customers can be visited more than once by the truck if it is convenient for the drone launch and recovery; the launch and reunion locations of the drone and truck can coincide; the drone's autonomy is considered unlimited; and the launch and reunion times can be considered insignificant.

According to the definition presented in [16], despite the author calling the approach TSP-D, premises of the FSTSP (no revisiting of customer nodes or launch and reunion at the same node) are used, and a Local Search algorithm and a metaheuristic GRASP (Greedy Randomized Adaptive Search Procedure) are employed. In a later work, the same authors used an improved hybrid Genetic Algorithm (GA) with Local Search procedure to minimize the total operational cost and delivery time. The algorithm provided optimal solutions for instances of up to one hundred customers [17].

In this work, the problem of optimizing the delivery service by a truck-drone system will be approached according to the definitions established for the FSTSP in [11], but with the mathematical formulation of the problem in MILP based on [18], which allowed the implementation of an exact model with only thirteen constraints (unlike the model proposed in [11], which contain 28 restrictions).

## III. PROBLEM DEFINITION AND MATHEMATICAL FORMULATION

In the problem, the initial node, defined as 0, and the final node, $c+1$, are composed of the same depot, and it is intended that the set of ten customers $C = \{1,..., c\}$, with $c = 10$ be served by the truck or drone. The positions of the customers are previously known. All ten customers in the set $C$ can be served by the truck, but only the customers in the subset $C' \subseteq C$ can be served by the drone.

The problem is constructed in the directed graph $G = (N, A)$, where: $N = \{0, 1,..., c+1\}$ contains all possible nodes; $N_0 = \{0, 1,..., c\}$ represents the possible initial nodes of an arc; and; and $N_+ = \{1,..., c+1\}$, represents the possible termination nodes of an arc. Thus, let A be the set of all arcs $(i, j)$, where $i \in N_0$ and $j \in N_+$, with $i \neq j$. Each arc $(i, j)$ is associated with two non-negative travel times (in minutes), $\tau_{ij}^C$ e $\tau_{ij}^D$, representing the travel time of that arc by the truck and drone, respectively.

The travel time matrices for the drone and truck are usually different, and the travel time between nodes 0 and $c+1$ is defined as 0, as they represent the same physical point (depot). The times spent servicing the customers, both for the drone and truck, are included in their respective travel times. A preparation time for the drone launch, given by $\sigma^L$, and a recovery time (landing on the truck), given by $\sigma^R$. Both actions are performed by the truck driver, who remains parked at a customer node during these procedures. However, the drone could also be launched from the depot. In this situation, $\sigma^L$ is equal to zero. The parameter $E$ represents the drone's autonomy, in minutes, restricting its use during sorties. The recovery time $\sigma^R$ contributes to the total calculation of autonomy, while $\sigma^L$ does not affect it, since the drone remains on the truck until takeoff. Thus, only $\sigma^L$ will contribute to the total delivery time.

A sortie is formally defined as $\langle i, j, k \rangle$, com $i \neq j \neq k$, where: $i \in N_0$ is the launch node; $j \in C'$ represents the customer to be served; and $k \in N_+$, is the node where the drone and truck reunite. Let $F$ be the set of all sorties that can be performed in accordance with the autonomy $E$ (for which the condition $\tau_{ij}^D + \tau_{kj}^D + \sigma^R \leq E$). The drone cannot be launched from the depot until the truck begins its route, and after each sortie, the battery is replaced or recharged for a new sortie. A sortie $\langle i, j, k \rangle$ may also become unfeasible if the truck exceeds the drone's autonomy $E$ when traversing the arc $(i, k)$.

Finally, the truck can serve customers during a drone sortie. For the reunion, synchronization is necessary, in which the vehicle that arrives first at a reunion point must wait for the other.

After detailing the sets and parameters of the problem, it is necessary to define the decision variables. Let $x_{ij} \in \{0, 1\}$ be equal to 1 if node $j \in N_+$ is visited after node $i \in N_0$, with $i \neq j$; 0 otherwise. The drone sorties are represented by the triple-index variable $y_{ijk} \in \{0, 1\}$, with $\langle i, j, k \rangle \in F$, equal to 1 if the sortie is performed, 0 otherwise. The binary Variable $z_i \in \{0,1\}$ equals 1 if the drone is on the truck; 0 otherwise. The non-negative variable $w_i$ represents the waiting time of the truck at node $i \in N$, while the non-negative Variable $t_i$, with $i \in N$, used to represent the synchronization time between the truck and the drone, including any waiting time of the drone for the truck.

The objective function (1) aims to minimize the arrival time of the truck and the drone at the final depot, where the drone may return to the depot before the truck. When this happens, the time difference will be considered by the wait time variable $w_{c+1}$. The delivery completion time can be decomposed into the following components: the total travel time of the truck; the total time required for drone launches and recoveries during the sorties performed; and the total time the truck waits for the drone.

$$\min \sum_{(i,j) \in A} \tau_{ij}^C x_{ij} + \sigma^R \sum_{\langle 0,j,k \rangle \in F} y_{0jk} + (\sigma^L + \sigma^R) \sum_{\langle i,j,k \rangle \in F, i \neq 0} y_{ijk} + \sum_{i \in N_+} w_i \quad (1)$$

Constraints (2) and (3) require that each customer be serviced only once by one of the two vehicles. In (4), the need for the truck to start and finish its route at the depot is imposed, while (5) defines the need to conserve the flow of the truck by combining (2) and (3).

$$\sum_{(i,j) \in A} x_{ij} + \sum_{\langle i,j,k \rangle \in F} y_{ijk} = 1, \text{para } j \in C \quad (2)$$

$$\sum_{(j,i) \in A} x_{ji} + \sum_{\langle i,j,k \rangle \in F} y_{ijk} = 1, \text{para } j \in C \quad (3)$$

$$\sum_{j \in N_+} x_{0j} = \sum_{i \in N_0} x_{i,c+1} = 1 \quad (4)$$

$$\sum_{(i,j) \in A} x_{ij} = \sum_{(j,i) \in A} x_{ji} \quad (5)$$

A series of constraints were established to ensure timing relationships. In these constraints, the constant $M$ represents a large value. Constraint (6) ensures that if the arc $(i,j)$ is traversed by the truck, then the time at node $j$ will be greater than or equal to the time spent up to node $i$, plus the time spent on the arc itself. Constraint (7) states that if a sortie $\langle i,k,j \rangle$, exists, then the time $t_j$ must be greater than or equal to the time spent by the drone on the sortie. Constraints (6) e (7) also prevent node $j$ from being visited by the truck before node $i$. Constraint (8) establishes the waiting time for the truck if it arrives at a rendezvous node $j$ before the drone. If the drone arrives at node j before the truck, the wait will be performed in hovering flight, and this time will be absorbed by $t_j$ in constraint (6), with the time included in the objective function.

$$t_j \geq t_i + \tau_{ij}^C - M(1 - x_{ij}), \text{para } (i,j) \in A \quad (6)$$

$$t_j \geq t_i + \tau_{ik}^D + \tau_{kj}^D - M(1 - \sum_{\langle i,k,j \rangle \in F} y_{ikj}), (i,j) \in A \quad (7)$$

$$w_j \geq t_j - t_i - \tau_{ij}^C - M(1 - x_{ij}), \text{para } (i,j) \in A \quad (8)$$

In (9), if a sortie $\langle i,j,k \rangle$ is performed, the time spent from drone launch to recovery must respect the drone's autonomy $E$.

$$t_k - t_i + \sigma^R - M(1 - \sum_{\langle i,j,k \rangle \in F} y_{ijk}) \leq E, \text{para } (i,k) \in A \quad (9)$$

Constraints (10) to (12) comprise relationships between the variable $z_i$, which defines whether a drone is aboard the truck or not, and $x_{ij}$ and $y_{ijk}$. (10) shows that the drone can be on the truck at node $i$. (11) attests that a sortie can start at node $i$ if the variable $z_i$ is equal to a 1, while (12) regulates the variable $z_i$, defining the presence of the drone on the truck during the vehicle's journey. Therefore, given $z_i = 1$ due to (11), (12) imposes $z_j = 0$, while if the sortie initiated at $i$ returns at $j$, $z_j$ may assume the value 1, which means that a new sortie could start at $j$. Returning to the general case where the sortie is not completed at $j$, the variables $z$ along the path (after the sortie is launched at $i$ and before the rendezvous at $l$) remain with the value 0, imposing that no other sortie can start until the truck reaches the rendezvous node $l$. Finally, (13) defines the start time of the journey, at the depot, as 0.

$$z_i \leq \sum_{(j,i) \in A} x_{ji}, \text{para } i \in N_+ \quad (10)$$

$$\sum_{\langle i,j,k \rangle \in F} y_{ijk} \leq z_i, \text{para } i \in N_0 \quad (11)$$

$$z_j \leq z_i - x_{ij} + \sum_{\langle l,j,k \rangle \in F} y_{ljk} - \sum_{\langle i,k,l \rangle \in F} y_{ikl} + 1, \quad (i,j) \in A \quad (11)$$

$$t_0 = 0 \quad (13)$$

## IV. HYBRID META-HEURÍSTIC

The use of metaheuristic optimization algorithms for combinatorial problems such as TSP and its variations is common. However, it can be observed in state-of-the-art algorithms the need for adaptations in traditional metaheuristics, incorporating purely heuristic elements specific to the problem at hand. These solutions, sometimes identified as hybrids [19], tend to be less susceptible to local minima, presenting a higher computational cost than purely heuristic solutions, but not as high as exact solutions [20]. In this context, the study of hybrid metaheuristic solutions adapted for combinatorial problems such as FSTSP is justified as a search for intermediate alternatives between exact and heuristic solutions, both in terms of computational performance and optimality of the solutions that can be found.

Thus, this work presents a solution proposal for FSTSP based on a Genetic Algorithm, with customized steps for the problem and called Hybrid Genetic to Flying Sidekick (HGenFS). The proposal has similarities with the work presented in [17], however, different solutions were adopted for the local search and reconstruction phases, as well as a new chromosome modeling.

### A. Chromosome Definition

In this work, the proposed chromosome to represent the basic individual of FSTSP was defined to include both the sequence of deliveries (equivalent to a regular TSP), and the definition of the means to be used. This chromosome model differs from that proposed in [17] by considering the means that make the delivery, drone or truck, as part of the individual. In [17], the authors define the individual as the complete sequence of deliveries (without specifying the vehicle that performed it), with this factor recalculated by a specific function each time a new individual is created.

The proposed model seeks to define a more specific individual for FSTSP, reducing the computational cost to process each of them, as the chosen means are already part of the individual and do not need to be selected at each modification. Thus, only an analysis of the feasibility for the drone takeoff and landing points to serve the already specified consumers is demanded, which is done during the process of creating new populations.

Each individual was defined as two vectors, one with the numbers of all customers, ordered in the sequence of service, and another, of the same size, with binary values indicating whether the delivery for the customer of the same index is fulfilled by drone or truck. A representation of this model is presented in Figure 2, in which the upper vector represents the delivery sequence with 0 and 11 representing the depot, and the other numbers representing each customer. The lower vector represents the definition of which means would be used for each customer, with 1 for drone deliveries and 0 for truck deliveries. Additionally, for each delivery by drone, a pair is defined indicating the index relative to takeoff and landing.

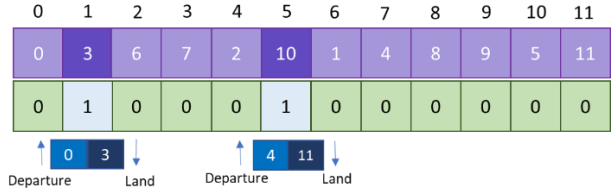

Fig. 2. Example of a chromosome that defines the individual.

*B. Crossover*

To create new individuals from the current population, parents are selected and merged in some way to generate potentially better offspring. This process is divided between selection and the merging itself, which is called crossover [17]. For selection, a method known as stochastic acceptance roulette was used, which, as discussed in [21], presents similar results to traditional roulette, but with a lower computational cost, O(1) versus O(log n).

Two crossover proposals were defined. The first consists of exchanging the different vectors between two parents, that is, each descendant is generated by adopting the delivery sequence of one parent and the means definition of the other. This proposal, quite simple, was called SWAP and evaluated for ease of implementation and low computational cost, given the chromosome model.

The second crossover proposal was a solution close to the method established in the literature for the TSP, known as OX (Ordered Crossover) [22]. However, due to the characteristic of the chromosome itself, the solution also incorporates information on drone deliveries, making it a specific solution called DX2. In this model, two cutting points are defined representing indices of the two vectors of the chromosome, delimiting a central region. The descendants are created by complementing the ends of one of the selected parents, sequentially with the missing customers, according to the order of the other parent. In this way, it is expected that the delivery sequences, including the choice of means (drone or truck), will be partially maintained, enabling the use of the good characteristics of both parents. It is worth noting that, after the creation of the offspring, it is necessary to recreate the drone route, defining takeoff and landing points, since previously defined points may no longer be available.

*C. Mutation*

As a form of population diversification, two mutation methods were adopted, with equal probability of occurrence of one or the other when a mutation is performed. The first method consisted of swapping the position of two customers, regardless of the delivery method used. Thus, the delivery sequence is modified, and it may also change for which customer a delivery defined as a drone is made, since the vector of selected means is not modified.

The second method consists of alternating the delivery method, that is, if the delivery of a randomly selected customer was made by drone, it switches to truck and vice versa. In practice, this mutation corresponds to a swap of the value of the means vector from 1 to 0, or vice versa. In this case, it is also necessary to reconstruct the drone route, if the swap is to add a new delivery by this means.

*D. Description of the Hybrid Genetic to Flying Sidekick (HGenFS)*

Generally, the HGenFS follows the steps of a standard genetic algorithm as presented in [23], with phases and characteristics adapted for the FSTSP that resulted in the algorithm presented in Table.

QUADRO I. PSEUDO-CODE DO ALGORITMO 1.

| Algorithm 1 - HGenFS |
|---|
| 1: Initialize Population |
| 2: **For** all *IND* in Population: |
| 3:     Eval(*IND*) |
| 4: **While** stopping criterion not met: |
| 5:     **For** all *N < nElite*: |
| 6:         Add individual *N* to new Population |
| 7:     **For** all *N >= nElite* and *N < nCrossOver/2*: |
| 8:         Select parents P1 e P2 |
| 9:         Obtain individuals *IND1* e *IND2* by *CrossOver* of P1 e P2 |
| 10:        **CreateRoute**(*IND1*) |
| 11:        **CreateRoute**(*IND2*) |
| 12:        Add IND1 e IND2 to new Population |
| 13:     **For** all *n >= nCrossOver*: |
| 14:         Create new Random indivídual IND_R |
| 15:        **CreateRoute**(*IND_R*) |
| 16:     **For** all *IND* in Population: |
| 17:        **Eval**(*IND*) |
| 18:     **For** all *IND* in evaluated Population: |
| 19:        **While** number of iterations < *nSwapSearch* limit: |
| 20:           Randomly swap positions in IND's route |
| 21:           **CreateRoute**(*IND*) |
| 22:           **Eval**(*IND*) |
| 23:        **While** number of iterations < *nDroneSearch* limit: |
| 24:           Store IND in IND_BACK |
| 25:           Alternate delivery method for a client |
| 26:           **CreateRoute**(*IND*) |
| 27:           **Eval**(*IND*) |
| 28:           **IF** quality of *IND* is worse than *IND_BACK*: |
| 29:              IND receive IND_BACK |

During execution, the Eval function is responsible for calculating the quality of the individual, represented by the total time to complete deliveries (objective function), with penalties for exceeding the drone's autonomy. Additionally, the CreateRoute function is responsible for finding takeoff and landing points to enable delivery by drone to a customer, points that, according to FSTSP, must correspond to customers served by truck. In this function, selection can be simply random among the possible points or by means of additional local search (selecting the best possible route), at a higher computational cost and potentially reducing diversity, and should be adopted according to the problem under study. If there are no satisfactory points, the delivery by drone is canceled, and the customer is served by truck.

HGenFS initializes with the creation of a new population (line 1), which for the tests performed was created randomly. The creation is done by initializing a vector with a random sequence of deliveries, as well as a randomly chosen delivery method, 1 or 0, in the second vector. The quality of each individual is obtained by calling the Eval function. Next, the algorithm enters the evolution loop until some stopping criterion is reached.

The first step in the process is the maintenance of a group called the elite for the next generation, defined by nElite (line 5). Then, descendants are created, two for each selected pair of parents P1 and P2 (lines 8 and 9). For these new individuals, routes for the drone are created (lines 10 and 11), that is, takeoff and landing points are selected using the CreateRoute function.

After creating the number of descendants equivalent to *nCrossOver*, if the population size has not yet been reached (a fixed size was adopted for all generations), random individuals are created to complete it (line 14). The new population is then evaluated, and local search is carried out according to the *nSwapSearch* and *nDroneSeach* parameters. Initially, a position inversion is made between two customers, a new drone route is created, and the individual is evaluated for the new quality. From this individual, for *nDroneSeach* times, the delivery method for one of the customers is randomly modified, a new route is created, and it is evaluated again (lines 20 to 22). If a better individual is obtained, it is kept; otherwise, the previous individual is restored. This process is repeated for *nSwapSearch* times.

After local search, individuals who improved by replacing the original ones in the population, the process is restarted, checking the stopping criterion and so on.

## V. RESULTS AND DISCUSSIONS

The mathematical models in PIM were solved in the AMPL environment [24], using the Gurobi solver (version 9.1.1), on a Lenovo Gaming 3i notebook with an Intel Core i7-10750H processor at 2.60 GHz and 8.00 Gb of RAM. The hybrid genetic algorithm was implemented in the C++ programming language, using a virtual machine with 40 Xeon Gold 6230R cores at 2.10 GHz and 128 Gb of RAM.

For both methods, nine instances (or data sets) out of a total of 36 proposals in [11], were used, each containing ten customers. This group was selected because it represents the scenarios with the greatest difference between the main works evaluated in these data [11], [17], [18], which would supposedly be the most challenging cases. In these scenarios, the customers were randomly distributed in an area of about thirteen square kilometers, while the depot was located near the center of gravity of the customers. The times defined for the drone were based on the Euclidean distance between the nodes, while the truck times were established according to the Manhattan distance [25].

In PIM, a maximum value of ten minutes was defined for each instance, so that the model would search for the ideal solution, while the ratio $|C'|/|C|$ was kept between 80% and 90% of the total number of customers, and the following parameters were adopted: $E = 20$ minutes; e $\sigma^L = \sigma^R = 1$ minute. No HGenFS, foram realizadas dez inicializações em paralelo, com um limite de 30 segundos. In HGenFS, ten parallel initializations were performed, with a limit of 30 seconds. Six groups of parameters were defined in order to demonstrate the effect of specific configurations of the proposed algorithm, as presented in Table I.

TABLE I. TESTS CASES FOR HGenFS.

|  | Case1 | Case2 | Case3 | Case4 | Case5 | Case6 |
|---|---|---|---|---|---|---|
| CrossOver | DX2 | DX2 | DX2 | SWAP | SWAP | SWAP |
| nSwapSearch | 10 | 1 | 20 | 10 | 1 | 20 |
| nDroneSearch | 10 | 20 | 1 | 10 | 20 | 1 |

The Table II presents, for each of the instances used (Instance column), the completion times of the routes (in minutes) obtained in [18] and in this work, using the exact (Exact column) and metaheuristic (Optimal column) methods. For HGenFS, the time in seconds that each Test Case took to find the optimal value is also shown, or 30.0 if the time limit was reached without obtaining that value.

TABLE II. COMPARISON BETWEEN THE RESULTS AND [18].

| Instance | Ref [18] (min) | Exact (min) | HGenFS | | | | | | |
|---|---|---|---|---|---|---|---|---|---|
|  |  |  | Optimal (min) | C1 (s) | C2 (s) | C3 (s) | C4 (s) | C5 (s) | C6 (s) |
| 437v6 | 48,604 | 48,604 | 48,604 | 3,5 | 1,9 | 0,9 | 4,1 | 0,7 | 3,4 |
| 437v12 | 56,849 | 56,849 | 56,849 | 5,0 | 0,6 | 5,8 | 14,7 | 1,1 | 13,1 |
| 440v6 | 44,506 | 44,506 | 44,506 | 22,1 | 9,7 | 8,2 | 4,6 | 5,5 | 9,2 |
| 440v7 | 49,900 | 49,900 | 49,900 | 0,7 | 4,3 | 9,9 | 16,4 | 4,5 | 7,0 |
| 440v8 | 62,700 | 62,700 | 62,700 | 11,5 | 13,4 | 2,7 | 27,7 | 30,0 | 16,2 |
| 440v9 | 42,533 | 42,533 | 42,533 | 6,5 | 2,6 | 30,0 | 30,0 | 7,5 | 30,0 |
| 443v7 | 65,523 | 65,523 | 65,523 | 25,9 | 17,6 | 14,9 | 30,0 | 15,3 | 30,0 |
| 443v10 | 47,935 | 47,935 | 47,935 | 2,5 | 3,6 | 28,5 | 10,4 | 11,7 | 17,2 |
| 443v11 | 57,382 | 57,382 | 57,382 | 17,2 | 2,8 | 1,8 | 10,4 | 1,5 | 14,3 |
| Average to achieve optimum (s) | | | | 10,5 | 6,3 | 11,4 | 16,5 | 8,6 | 14,3 |

From the results presented in Table II, it is possible to observe that the values found by the exact method coincided with the values of Dell'Amico, Montemanni, and Novelanni [18], with the optimal solution being obtained in all cases with a maximum processing time of 98 seconds and an average of 40.6 seconds. These results confirmed that the proposal presented by the authors, and implemented for this work, can find optimal solutions for the FSTSP with ten customers in a satisfactory time, even with a common personal computer.

HGenFS, with the settings of cases 1 and 2, was able to find the optimal results for all instances within the 30-second limit. For the other cases, exceptions are highlighted in red in the table, as the limit was reached without obtaining the optimal solution. Comparing the different configurations for the proposed algorithm, regarding the crossover method, it can be observed that DX2 was more efficient (cases 1 to 3), always reaching the optimal solution in less time than SWAP (cases 4 to 6) for the same local search configurations. It is worth noting that in the three configurations using SWAP, there were cases where the optimal solution was not reached within 30 seconds.

Regarding local search, it can be observed that prioritizing the search in the selection of means (*nDroneSearch*) significantly improved the results, regardless of the crossover used. For example, in Case 1, using DX2 crossover and a balanced search (*nDroneSearch* = 10, *nSwapSearch* = 10), the average time until obtaining the optimal solution was 10.5 seconds, while in Case 2, modifying *nDroneSearch* to 20 and *nSwapSearch* to 1, we obtained 6.3 seconds. A similar gain can also be observed with the SWAP crossover in Case 5. In the opposite sense, in Cases 3 and 6, it is observed that prioritizing the search for better delivery sequences made HGenFS less efficient, including generating the only case where the optimal solution was not obtained with the DX2 crossover (test 440v9 in Case 3).

## VI. CONCLUSION

By means of the exact solution implemented, it was possible to confirm the results found in [18], demonstrating the feasibility of solving the FSTSP proposal for up to ten customers in seconds, while the solution presented in [11] was unable to solve even after thirty minutes of processing.

The proposed hybrid algorithm, called HGenFS, also proved to be quite efficient in solving the FSTSP, finding the optimal solutions in all selected tests in an average time of 6.27s (with the DX2 crossover model) and prioritizing local search to find the best delivery means at the end of each generation.

Although the results already demonstrate that HGenFS can significantly outperform exact solutions in the state of the art for ten customers, the algorithm needs to be tested with more complex problems, allowing for a more direct comparison with other hybrid and/or heuristic solutions. However, the results found already demonstrate that the proposed solution is effective and presents itself as an alternative to existing literature to solve the FSTSP.

This Paper is a translation of [26] by the same authors, and it aims to provide a comprehensive understanding of the Flying Sidekick TSP (FSTSP) problem and the proposed hybrid Genetic Algorithm (HGenFS) for its optimization. While the present work focuses on solving FSTSP with up to ten clients, the authors plan to conduct future tests with a larger number of clients and multiple drones, and provide updates accordingly.